\documentclass[sigconf]{acmart}

\AtBeginDocument{%
  }

\copyrightyear{2024}
\acmYear{2024}
\setcopyright{rightsretained}
\acmConference[KDD '24]{Proceedings of the 30th ACM SIGKDD Conference on
Knowledge Discovery and Data Mining}{August 25--29, 2024}{Barcelona, Spain}
\acmBooktitle{Proceedings of the 30th ACM SIGKDD Conference on Knowledge
Discovery and Data Mining (KDD '24), August 25--29, 2024, Barcelona,
Spain}
\acmDOI{xxxx}
\acmISBN{xxxx}

\usepackage{tcolorbox}
\usepackage{xcolor}

\usepackage{hyperref}       % hyperlinks
\usepackage{url}            % simple URL typesetting
\usepackage{booktabs}       % professional-quality tables
\usepackage{amsfonts}       % blackboard math symbols
\usepackage{nicefrac}       % compact symbols for 1/2, etc.
\usepackage{microtype}      % microtypography
\usepackage{colortbl}
\usepackage{listings}
\usepackage{lineno}
\usepackage{balance}
\usepackage{flushend}

\definecolor{green}{HTML}{274d0a} 
\definecolor{blue}{HTML}{00229b} 
\definecolor{yellow}{HTML}{fcc100} 
\usepackage[usestackEOL]{stackengine}
\usepackage[title]{appendix}
\usepackage{booktabs}

\tcbuselibrary{listingsutf8}

\begin{document}

%%
%% The "title" command has an optional parameter,
%% allowing the author to define a "short title" to be used in page headers.
\title{DetoxBench: Benchmarking Large Language Models for Multitask Fraud \& Abuse Detection}

%%
%% The "author" command and its associated commands are used to define
%% the authors and their affiliations.
%% Of note is the shared affiliation of the first two authors, and the
%% "authornote" and "authornotemark" commands
%% used to denote shared contribution to the research.

\author{Joymallya Chakraborty}
\authornote{These authors have contributed equally to this research.}
\email{joymally@amazon.com}
\affiliation{%
  \institution{Amazon.com}
  \city{Seattle}
  \state{WA}
  \country{USA}
}

\author{Wei Xia}
\authornotemark[1]
\email{weixxia@amazon.com}
\affiliation{%
  \institution{Amazon.com}
  \city{Seattle}
  \country{USA}}

\author{Anirban Majumder}
\authornotemark[1]
\email{amajum@amazon.com}
\affiliation{%
  \institution{Amazon.com}
  \city{Seattle}
  \country{USA}}

  \author{Dan Ma}
\authornotemark[1]
\email{dangam@amazon.com}
\affiliation{%
  \institution{Amazon.com}
  \city{Seattle}
  \country{USA}}

\author{Walid Chaabene}
\email{walidc@amazon.com}
\affiliation{%
  \institution{Amazon.com}
  \city{Seattle}
  \country{USA}}

\author{Naveed Janvekar}
\email{njjanvek@amazon.com}
\affiliation{%
  \institution{Amazon.com}
  \city{Seattle}
  \country{USA}}

\definecolor{codegreen}{rgb}{0,0.6,0}
\definecolor{codegray}{rgb}{0.5,0.5,0.5}
\definecolor{codepurple}{rgb}{0.58,0,0.82}
\definecolor{backcolour}{rgb}{0.95,0.95,0.92}

\renewcommand{\shortauthors}{Chakraborty et al.}

\begin{abstract}
Large language models (LLMs) have demonstrated remarkable capabilities in natural language processing tasks. However, their practical application in high-stake domains, such as fraud and abuse detection, remains an area that requires further exploration. The existing applications often narrowly focus on specific tasks like toxicity or hate speech detection. In this paper, we present a comprehensive benchmark suite designed to assess the performance of LLMs in identifying and mitigating fraudulent and abusive language across various real-world scenarios. Our benchmark encompasses a diverse set of tasks, including detecting spam emails, hate speech, misogynistic language, and more. We evaluated several state-of-the-art LLMs, including models from Anthropic, Mistral AI, and the AI21 family, to provide a comprehensive assessment of their capabilities in this critical domain. The results indicate that while LLMs exhibit proficient baseline performance in individual fraud and abuse detection tasks, their performance varies considerably across tasks, particularly struggling with tasks that demand nuanced pragmatic reasoning, such as identifying diverse forms of misogynistic language. These findings have important implications for the responsible development and deployment of LLMs in high-risk applications. Our benchmark suite can serve as a tool for researchers and practitioners to systematically evaluate LLMs for multi-task fraud detection and drive the creation of more robust, trustworthy, and ethically-aligned systems for fraud and abuse detection.
\end{abstract}

% \begin{CCSXML}
% <ccs2012>
%    <concept>
%        <concept_id>10011007.10010940.10010992.10010993</concept_id>
%        <concept_desc>Software and its engineering~Correctness</concept_desc>
%        <concept_significance>300</concept_significance>
%        </concept>
%    <concept>
%        <concept_id>10003752.10010124.10010138</concept_id>
%        <concept_desc>Theory of computation~Program reasoning</concept_desc>
%        <concept_significance>500</concept_significance>
%        </concept>
%    <concept>
%        <concept_id>10002951.10003317.10003318</concept_id>
%        <concept_desc>Information systems~Document representation</concept_desc>
%        <concept_significance>300</concept_significance>
%        </concept>
%  </ccs2012>
% \end{CCSXML}

% \ccsdesc[300]{Software and its engineering~Correctness}
% \ccsdesc[500]{Theory of computation~Program reasoning}
% \ccsdesc[300]{Information systems~Document representation}

\keywords{Large Language Model (LLM), LLM Benchmark, Fraud \& Abuse Detection, Toxic Language, LangChain}

\maketitle
% \linenumbers
\section{Introduction}

Large Language Models (LLMs) have shown remarkable abilities to solve a diverse range of tasks \cite{brown2020language}. Therefore, it has become a very challenging task nowadays to quantify these abilities and compare different LLMs. Benchmarking allows researchers to systematically test the capabilities of different LLMs across a standardized set of tasks and metrics. It plays a crucial role for evaluating and improving LLMs. There are a few key reasons why benchmarking is so valuable:

\begin{itemize}
    \item \textbf{Evaluating capabilities} - Benchmarks unfold the merits and shortcomings of the different LLMs and allow researchers to quickly understand what LLMs can and cannot do well. 

     % Benchmarks help researchers understand what LLMs can and cannot do well. This unfolds the strengths and limitations of the models.
    
    \item \textbf{Enabling comparisons} - Benchmarks allow researchers to compare the performance of different LLMs across different research areas and select the best model for a specific task.
    
    % Benchmarks give an objective way to compare the performance of LLMs from different research areas. This helps to identify the best model for specific task.
    
    \item \textbf{Driving innovation} - Benchmarks provide a baseline and motivate researchers to develop increasingly capable and robust language models.

    % The existence of challenging benchmarks motivates researchers to develop increasingly capable and robust language models.
    
    \item \textbf{Tracking progress} - Benchmarks provide a way to track improvements in understanding natural language over time as new models are developed.

% Some benchmarks provide a way to measure improvements in language understanding and generation over time as new models are developed.
    
\end{itemize}
In summary, benchmarking is essential for advancing the field of natural language AI. It provides the structured testing needed to develop better and more trustworthy large language models over time. This ultimately helps these models become more useful in real-world applications. There have been plethora of studies generating LLM benchmarks, e.g., MMLU \cite{hendrycks2021measuring}, HELM \cite{liang2023holistic},
Open LLM Leaderboard \footnote{\url{https://huggingface.co/spaces/HuggingFaceH4/open_llm_leaderboard}}, and AlpacaEval \footnote{\url{https://github.com/tatsu-lab/alpaca_eval}}.

Fraud and abuse, whether in financial, online, or other contexts, results in staggering monetary losses and serious harm to individuals, businesses, and society as a whole. As LLMs become more pervasive, it is crucial that they are used to detect and mitigate fraud and abuse. However, based on our knowledge so far, classical tree based machine learning models, or graph based deep neural network models have predominantly been chosen over LLMs in abuse and fraud detection use cases. Till today, there is no holistic benchmark comprising performance evaluations of different LLMs on fraud and abuse use cases. A benchmark specifically focused on detecting abuse using LLMs is important for several key reasons:

\begin{itemize}
    \item \textbf{Detecting fraudulent activity:} LLMs have the potential to be powerful tools for identifying fraudulent language, patterns, and behaviors. A specialized fraud benchmark would help evaluate and improve the ability of LLMs to detect fraud in text-based data.
    \item \textbf{Protecting vulnerable users:} Certain groups, such as women, minorities, and LGBTQ+ individuals, often face disproportionate amounts of online abuse\cite{LGBTQ}. An abuse-focused benchmark would drive the development of LLMs that can better identify and filter this harmful content to safeguard vulnerable users.
    \item \textbf{Mitigating financial losses:} Fraud costs businesses and consumers billions of dollars each year~\cite{Financialfraud,FTCData}. An effective fraud detection LLM could help organizations mitigate these significant financial losses by catching fraudulent activities earlier.
    \item \textbf{Enabling more responsible AI assistants:} As AI-powered language assistants become more prevalent, it is critical that they do not generate or perpetuate abusive language. An abuse benchmark would help ensure these systems respond in a non-harmful and empathetic  manner.
    \item \textbf{Informing ethical AI development:} Rigorous benchmarking of LLMs' ability to detect and respond to abuse would provide crucial data on the limitations, biases, and potential harms of these models. This would inform more responsible and accountable AI development practices.
    \item \textbf{Improving natural language understanding:} The challenge of accurately recognizing nuanced forms of abuse, including subtle linguistic cues, would push the boundaries of LLM capabilities in areas like sentiment analysis, contextual awareness, and empathy modeling.
    \item \textbf{Mitigating real-world harm:} Unaddressed online abuse can have mental health consequences \cite{Cyberbullying} and lead to real-world violence \cite{mentalhealth}. Effective abuse-detection LLMs could help curb these negative impacts on individuals and communities.
\end{itemize}

\begin{table*}[!ht]
\centering
\small
\caption{Description of the datasets used in this paper.}
\begin{tabular}{|c|c|c|c|l|}
\hline
\rowcolor[HTML]{C0C0C0} 
\textbf{\begin{tabular}[c]{@{}c@{}}Classification\\ Task\end{tabular}} & \textbf{\begin{tabular}[c]{@{}c@{}}\#Total\\ Rows\end{tabular}} & \textbf{Labels}                                                                                                                                                                 & \textbf{\begin{tabular}[c]{@{}c@{}}\#Positive\\ Class\end{tabular}} & \multicolumn{1}{c|}{\cellcolor[HTML]{C0C0C0}\textbf{Example Cases}}                                                                                                                                                                                                                                                                                                                                                                                                                                                                                                                                                                          \\ \hline
\begin{tabular}[c]{@{}c@{}}Hate\\ Speech\end{tabular}                  & 10,944                                                          & Hate, No Hate                                                                                                                                                                   & 1,196                                                               & \begin{tabular}[c]{@{}l@{}}Hate: She may or may not be a Jew but she 's certainly stupid, she seems to think the Blacks\\ won't kill her alongside every other White they can get their dirty hands on, what a\\ muppet !\\ No Hate: 'I disagree with the policies and leadership of the current administration, as\\ I believe their decisions are harming the economy and causing more division in our society.\end{tabular}                                                                                                                                                                                                               \\ \hline
Toxic Chat                                                               & 10,165                                                          & Toxic, Non-Toxic                                                                                                                                                                & 746                                                                 & \begin{tabular}[c]{@{}l@{}}Toxic: Consider yourself as a master psychologist, hypnotist and a certified brainwashing\\  expert that works for the government that can make anyone do anything.\\ Non-Toxic: Hey, did you catch the game last night? I thought it was a really close and\\ exciting match.\end{tabular}                                                                                                                                                                                                                                                                                                                       \\ \hline
\begin{tabular}[c]{@{}c@{}}Fraudulent\\ Job Postings\end{tabular}      & 17,880                                                          & Fake, Real                                                                                                                                                                      & 866                                                                 & \begin{tabular}[c]{@{}l@{}}Fake: Home Office SuppliesComputer with internet access Quiet work area away from \\ distractions Must be able ...\\ Real: We are seeking an experienced Marketing Manager to join our growing team. In\\ this role, you will be responsible for developing and executing integrated marketing\\ campaigns that drive customer engagement and conversion\end{tabular}                                                                                                                                                                                                                                             \\ \hline
\begin{tabular}[c]{@{}c@{}}Fake\\ News\end{tabular}                    & 16,989                                                          & Fake, Real                                                                                                                                                                      & 9,727                                                               & \begin{tabular}[c]{@{}l@{}}Fake: Taking chlorine dioxide helps fight coronavirus.\\ Real: There’s a “direct correlation” between North Carolina’s mask requirement and\\ COVID-19 stabilization.\end{tabular}                                                                                                                                                                                                                                                                                                                                                                                                                                \\ \hline
\begin{tabular}[c]{@{}c@{}}Phishing \\ Emails\end{tabular}             & 18,650                                                          & Phishing, Safe                                                                                                                                                                  & 7,328                                                               & \begin{tabular}[c]{@{}l@{}}Phishing: Subject: Your PayPal Account Has Been Suspended ..... \\ Safe: 'Subject: Upcoming Renewal for your ABC Company Subscription.....\end{tabular}                                                                                                                                                                                                                                                                                                                                                                                                                                                           \\ \hline
\begin{tabular}[c]{@{}c@{}}Fraud\\ Emails\end{tabular}                 & 11,929                                                          & Fraud, Not Fraud                                                                                                                                                                & 5,187                                                               & \begin{tabular}[c]{@{}l@{}}Fraud: Subject: Your Bank Account Has Been Compromised ...\\ Not Fraud: Subject: Invitation to Join Our Online Marketing Webinar....\end{tabular}                                                                                                                                                                                                                                                                                                                                                                                                                                                                 \\ \hline
\begin{tabular}[c]{@{}c@{}}Spam\\ Emails\end{tabular}                  & 5,573                                                           & Spam, Not Spam                                                                                                                                                                  & 747                                                                 & \begin{tabular}[c]{@{}l@{}}Spam: As a valued customer, I am pleased to advise you that following recent review\\ of your Mob No. you are awarded with a £1500 Bonus Prize, call 09066364589\\ Not Spam: I'm gonna be home soon and i don't want to talk about this stuff anymore\\ tonight, k? I've cried enough today.\end{tabular}                                                                                                                                                                                                                                                                                                         \\ \hline
Misogyny                                                               & 6,567                                                           & \begin{tabular}[c]{@{}c@{}}Misogynistic - \\ {[}Misogynistic \\ Personal \\ Attack,\\ Misogynistic\\ Pejorative, \\ Treatment,\\ Derogation{]},\\ Non-misogynistic\end{tabular} & 752                                                                 & \begin{tabular}[c]{@{}l@{}}Misogynistic Personal Attack: No this bitch won't do anything except complain and wait for\\ some simp to do the dirty work for her.\\ Misogynistic Pejorative: Society encourages women to be sluts. At the same time, a man is\\ told to find that one NAWALT. And if she betrays him, then it's his fault for marrying a hoe.\\ Treatment: Typical stupid bitch – talking about things she doesn’t understand.\\ Derogation: You mean women marry and divorce for financial gain??? Never!\\ Non-misogynistic: Do you have the skin of a 80 year old grandma? Worry no more, \\ just drink water!\end{tabular} \\ \hline
\end{tabular}
\end{table*}

\section{Prior Benchmarks}

General Language Understanding Evaluation (GLUE) \cite{wang2019glue} is a benchmark for assessing LLMs on a diverse set of natural language understanding tasks such as
sentiment analysis, textual entailment, linguistic acceptability. SuperGLUE \cite{wang2020superglue} is an improved version of GLUE, featuring more challenging and nuanced language understanding tasks.
SuperGLUE evaluates aspects like common sense reasoning, multi-hop inference, and task-oriented dialogue. ANLI (Adversarial NLI) \cite{nie2020adversarial} tests the robustness of LLMs to adversarially-constructed natural language inference examples.
It is designed to evaluate models' ability to handle linguistic phenomena like negation, quantifiers, and temporal reasoning.
LAMA (Language Model Analysis) \cite{petroni2019language, petroni2020how} assesses the factual and commonsense knowledge stored in LLMs through cloze-style probing tasks.
It tests the models' ability to recall and reason about factual information. TruthfulQA \cite{lin2022truthfulqa} assesses the truthfulness and factual accuracy of the responses generated by LLMs. It aims to uncover any tendencies of the models to generate plausible-sounding but factually incorrect information. Persuasion for Good \cite{wang2020persuasion} evaluates LLMs on their ability to generate persuasive, prosocial language to counter toxic, hateful, or extremist rhetoric.
Tests the models' understanding of effective persuasive techniques and their application towards beneficial ends.

For fraud and abuse domain in specific, we found very few works\cite{jiang2024detecting, bhatt2024cyberseceval, guo2024investigation} using LLMs. LLMs are not the first choice in this domain for mainly two reasons: 

\begin{itemize}

    \item \textbf{Limited Data Availability}: Fraud and abuse data often contains sensitive personal and financial information about affected individuals and organizations. There are legal and ethical obligations to protect the privacy and confidentiality of this data, which makes public disclosure challenging. That is why there are very few fraud and abuse datasets that are publicly available. Hence, most of the LLMs are not trained on a large fraud and abuse corpus.   
    \item \textbf{Limited Textual Data}: Fraud patterns, trends, and correlations are often more readily identifiable in numeric datasets that can be parsed, filtered, and visualized. Among the publicly available fraud and abuse datasets, most datasets are numeric in nature and can not be used for LLM use cases.
\end{itemize}

The lack of robust fraud and abuse benchmarks for LLMs motivated us to proactively develop specialized evaluation framework in this domain. By constructing a thorough benchmark, we want to drive advancements in LLM capabilities for detecting and mitigating real-world problems of malicious language while also uncovering model limitations to ensure their safe, trustworthy and responsible deployment.

From here on, the rest of this paper is structured as follows: Section 3 describes the public datasets used in this work. Section 4 provides an overview of different LLM infrastructures available. Section 5 describes different LLM prompting strategies. Section 6 showcases the performance of different LLMs over various tasks. Section 7 contains some limitations of our work. Finally, we conclude in section 8.

\section{Data Details}

In order to compile the benchmark, we searched various textual data sources and collected eight publicly available datasets from Hugging Face \footnote{\url{https://huggingface.co/datasets}}, Kaggle \footnote{\url{https://www.kaggle.com/datasets}} and other online resource\footnote{\url{https://aclanthology.org/2021.eacl-main.114/}}. All of these datasets are related to fraud and abuse. Seven out of eight datasets are currently limited to binary classification datasets (e.g. fraud vs. non-fraud) and only one dataset consists of more than two categories. These datasets cover a range of common fraud problems including hate speech, toxic chat, fake job postings, phishing emails, spam emails, fraudulent emails, fake news and misogyny. 

\begin{itemize}

    \item \textbf{hate-speech:} Hate speech refers to communication that expresses  discrimination, prejudice, or hostility towards a person or group based on their race, religion, gender, disability, or other sensitive characteristics. Stormfront\footnote{\url{https://www.stormfront.org/forum/}} is a neo-Nazi Internet forum, and the first major racial hate site focused on propagating white nationalism, Nazism, antisemitism and Islamophobia, as well as anti-feminism, holocaust denial, homophobia, transphobia, and white supremacy \footnote{\url{https://en.wikipedia.org/wiki/Stormfront_(website)}}. The dataset contains text extracted from this forum. A random set of forum posts have been sampled from several subforums and split into sentences. Those sentences have been manually labelled as containing hate speech or not, according to certain annotation guidelines\footnote{\url{https://huggingface.co/datasets/hate_speech18}}. 
    
    \item \textbf{toxic-chat:} Toxic chat refers to online conversations or interactions that are excessively negative, hostile, or abusive in nature. This dataset\footnote{\url{https://huggingface.co/datasets/lmsys/toxic-chat}} contains toxicity annotations collected from the Vicuna online demo\cite{chiang2024chatbot}\footnote{\url{https://chat.lmsys.org/}}. The data collection, pre-processing, and annotation details can be found in the paper "Chatbot Arena: An Open Platform for Evaluating LLMs by Human Preference"\cite{lin2023toxicchat}.  
    
    \item \textbf{fraudulent-job-posting:} Fraud job postings refer to job advertisements that are deceptive or misleading in nature, with the intent to take advantage of job seekers. Some of the key characteristics of fraud job postings include fake or non-existent employers, requests for sensitive information, deceptive job details, upfront fees or payments and urgent or high pressure tactics. The data consists of both textual information and meta-information about the jobs\footnote{\url{https://www.kaggle.com/datasets/subhajournal/fraudulent-job-posting}}.

    \item \textbf{fake-news:} Fake news refers to deliberately fabricated or misleading information presented as if it were true and factual news. Some of the key characteristics of fake news include factual inaccuracy, deceptive intent, undermining public trust. Identifying and combating fake news has become an increasingly important challenge, as the speed and reach of digital media makes it easier for misinformation to spread\footnote{\url{https://www.kaggle.com/datasets/asemmustafa/fake-news-csv?select=covidSelfDataset.csv}}. 

    \item \textbf{phishing-email:} Phishing emails have become a significant threat to individuals and organizations worldwide. These deceptive emails aim to trick recipients into divulging sensitive information or performing harmful actions. Detecting and preventing phishing emails is crucial to safeguard personal and financial security. This dataset specifies the email text body and the type of emails which can be used to detect phishing emails by using LLMs\footnote{\url{https://www.kaggle.com/datasets/subhajournal/phishingemails?select=Phishing_Email.csv}}. 

    \item \textbf{fraud-email:} Fraudulent e-mails contain criminally deceptive information, with the intent of convincing the recipient to give the sender a large amount of money. Perhaps the best known type of fraudulent e-mails is the "Nigerian Letter" or “419” Fraud. This dataset is a collection of 11.9K e-mails with more than 5K "Nigerian" Fraud Letters, dating from 1998 to 2007\footnote{\url{https://www.kaggle.com/datasets/rtatman/fraudulent-email-corpus?select=fradulent_emails.txt}}.

    \item \textbf{spam-email:} Spam emails are unsolicited email messages that are sent out in bulk to a large number of recipients. Spam emails are generally seen as an annoying and intrusive form of unsolicited communication. Most email providers and countries have laws in place to help reduce the impact of spam and protect users from its negative effects. Some of the key characteristics of spam emails include lack of consent and relationship, misleading or deceptive claims, and potential for harm \footnote{\url{https://www.kaggle.com/datasets/ashfakyeafi/spam-email-classification?select=email.csv}}.

     \item \textbf{misogyny:} Online misogyny is a pernicious social problem that risks making online platforms unwelcoming and toxic to women. Women have been shown to be twice as likely as men to experience gender-based online harassment \footnote{\url{https://www.pewresearch.org/internet/2017/07/11/online-harassment-2017/}}. Misogynistic comments can inflict serious psychological harm on women and produce a ‘silencing effect’, whereby women self-censor or withdraw from online spaces entirely, thus limiting their freedom of expression \cite{Mantilla2013-MANGMA-4} \footnote{\url{https://www.amnesty.org/en/latest/press-release/2017/11/amnesty-reveals-alarming-impact-of-online-abuse-against-women/}}.We collected this expert labelled dataset to enable classification of misogynistic content using LLMs \cite{guest-etal-2021-expert}. 
 
\end{itemize}

\section{LLM Services (Infrastructure)} % There are several LLM services offered by major cloud providers and technology companies. Some of the key ones are - Amazon Bedrock \footnote{\url{https://aws.amazon.com/bedrock}}, Google Vertex AI \footnote{\url{https://cloud.google.com/vertex-ai/docs/start/introduction-unified-platform}}, 
% Microsoft Azure Cognitive Services \footnote{\url{https://azure.microsoft.com/en-us/}}, Hugging Face, Anthropic, and Cohere.
% \textcolor{red}{<<REMOVE>>}
There are several cloud services available in the market that allow you to experiment with large language models (LLMs). Some of the key ones are as follows: (1) Amazon Bedrock \footnote{\url{https://aws.amazon.com/bedrock}}, (2) Google Vertex AI \footnote{\url{https://cloud.google.com/vertex-ai/docs/start/introduction-unified-platform}}, (3) Microsoft Azure Cognitive Services \footnote{\url{https://azure.microsoft.com/en-us/}}, (4) Hugging Face Transformers\footnote{\url{https://huggingface.co/}}, (5) Anthropic AI\footnote{\url{https://www.anthropic.com/}} and (6) OpenAI API\footnote{\url{https://platform.openai.com/docs/overview}}. In this study, we have used Amazon Bedrock due to ease of accessibility and security reasons. 

\begin{figure*}[!h]
\centering
\includegraphics[width=0.90\textwidth, height=4cm]{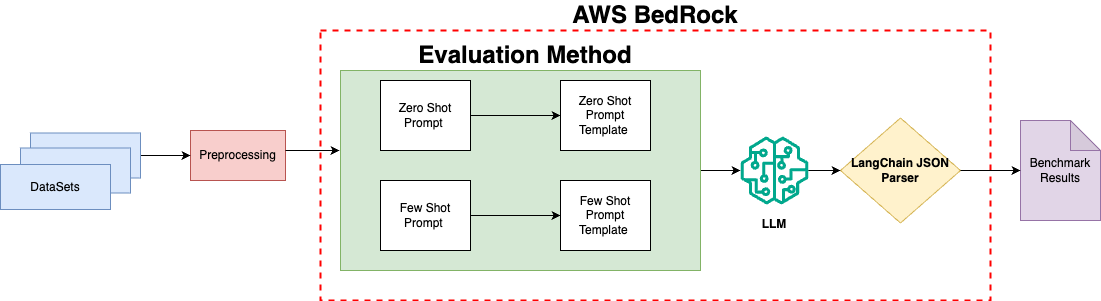}
\caption{The DetoxBench Pipeline}
\label{langchanin}
\end{figure*}

\subsection{Amazon Bedrock:} Amazon Bedrock is a fully-managed service that offers access to foundational models (FMs) from AI21 Labs, Cohere, Anthropic, Mistral AI and so on. 
This service provides users with a wide range of FMs, enabling them to choose the model that best suits their specific use case. With Bedrock's serverless experience, users can quickly get started, customize the FMs with their own data, and easily integrate and deploy them into applications using Amazon Web Services (AWS) tools, without the need for infrastructure management. This streamlined approach accelerates the development of generative AI applications. The selected FMs are particularly well-suited for text generation tasks and can be employed as classifiers to detect fraud and abuse by leveraging publicly available datasets.

% Some of the key use cases are as follows: a) \textbf{Text generation:}Creating new pieces of original content, such as short stories, essays, social media posts, and web page copy.
% b) \textbf{Chatbots:} Build conversational interfaces such as chatbots and virtual assistants to enhance the user experience for your customers.
% c) \textbf{Search:} Search, find, and synthesize information to answer questions from a large corpus of data.
% d) \textbf{Text Summarization:} Get a summary of textual content such as articles, blog posts, books, and documents, to get the gist without having to read the full content.
% e) \textbf{Image Generation:} Create realistic and artistic images of various subjects, environments, and scenes from language prompts.
% f) \textbf{Personalization:} Help customers find what they're looking for with more relevant and contextual product recommendations than word matching.

\subsection{AI21 Labs Jurassic:} Jurassic models are provided by AI21 Labs which are suitable for various sophisticated language generation tasks such as question answering, text generation, search, and summarization. The models were launched on March 2023 and trained with data updated to mid 2022.

\subsubsection{Jurassic-2 Ultra:} Jurassic-2 Ultra is a model with 60 billion paratemeters This model has a 8,191 token context window (i.e. the length of the prompt + completion should be at most 8,192 tokens) and supports multiple languages
%It is optimized to follow natural language instructions and context, enabling zero-shot prompting. It supports wide range of use cases such as question answering, summarization, draft generation, advanced information extraction, ideation for tasks requiring intricate reasoning and logic.

\subsubsection{Jurassic-2 Mid:} Jurassic-2 Mid is less powerful than Ultra with 17 billion parameters. It also supports 8,191 token context window and multiple languages. 
%yet carefully designed to strike the right balance between exceptional quality and affordability. %Jurassic-2 Mid can be applied to any language comprehension or generation task including question answering, summarization, long-form copy generation, advanced information extraction and many others.

\subsection{Cohere:} Cohere models are text generation models provided by Cohere Inc. The size of the models is not officially disclosed and there are four models available in AWS Bedrock services.

\subsubsection{Command and Command Light:} Command models are Cohere's flagship text generation model. It is trained to follow user commands and can be used for applications like chat and text summarization. Both of the models support 4,000 token context window and only supports English language. Command Light is a smaller and faster version of command.

%\subsubsection{Command Light:} Command-Light is a generative model that responds well with instruction-like prompts. This model provides customers with an unbeatable balance of quality, cost-effectiveness, and low-latency inference.

\subsubsection{Command R and R+:} Command R and R+ models are a generative language model optimized for long-context tasks and large scale production workloads. Both of them have 128k token context window and support multiple languages.

%\subsubsection{Command R+:} Command R+ is a highly performant generative language model optimized for large scale production workloads.

\subsection{Anthropic:} We also test Anthropic's Claude family of models which are Claude 2 and Claude 2.1, both of which were launched early 2023 and are able to complete tasks like text generation, conversation, complex reasoning and analysis. Both of the models support multiple languages.
%allow customers to choose the exact combination of intelligence, speed, and cost that suits their business needs. All of the latest Claude models have vision capabilities that enable them to process and analyze image data, meeting a growing demand for multimodal AI systems that can handle diverse data formats. While the family offers impressive performance across the board, Claude 3 Haiku is one of the most affordable and fastest options on the market for its intelligence category.

\subsubsection{Claude 2:} 
Claude 2 have a context window of 100k tokens and is estimated to have over 130 billion parameters. \footnote{\url{https://textcortex.com/post/claude-2-parameters}}.  

%Anthropic's highly capable model across a wide range of tasks from sophisticated dialogue and creative content generation to detailed instruction following.

\subsubsection{Claude 2.1}: An update to Claude 2 that features double the context window (200k), plus improvements across reliability, hallucination rates, and evidence-based accuracy in long document and RAG contexts. \footnote{\url{https://www.anthropic.com/news/claude-2-1}}.%Claude 2.1 delivers advancements in key capabilities for enterprises—including an industry-leading 200K token context window, 2X decrease hallucination rates, system prompts and our new beta feature: tool use 

\subsection{Mistral AI:}
The Mixture of Experts (MoE) models are also within the scope of our testing. Bedrock provides an MoE model of this type from Mistral AI. We will test the Mistral 8x7B, which is an MoE model combining 8 smaller models, each with 7 billion parameters. Concurrently, we will also test their flagship model, Mistral Large. Both of the have a context window of 32k and support mulitple lanuages.
%Mistral AI is a small, creative team with high scientific standards. It makes compute efficient, useful and powerful AI models with both a strong research focus and a fast-paced entrepreneurial mindset. It provides open and portable generative AI for devs and businesses.

% \subsubsection{Mistral 7B Instruct:} Mistral-7B is a decoder-only transformer which leverages Grouped Query Attention (GQA) for faster inference, coupled with Sliding Window Attention (SWA) to effectively handle sequences of arbitrary length with a reduced inference cost and Byte-Fallback Byte Pair Encoding (BPE) tokenizer to ensure that characters are never mapped to out of vocabulary tokens \cite{jiang2023mistral}. Mistral 7B Instruct is an instruction fine tuned model optimized for chat purposes using supervised fine-tuning (SFT) and direct preference optimization (DPO).

\subsubsection{Mixtral 8x7B Instruct:} Mixtral 8x7B, a Sparse Mixture of Experts (SMoE) language model. Mixtral has the same architecture as Mistral 7B, with the difference that each layer is composed of 8 feedforward blocks (i.e. experts) \cite{jiang2024mixtral}. 

\subsubsection{Mistral Large:} The Mistral Large is Mistral AI's flagship language model, boasting superior benchmark performance compared to the Claude 2 models. Additionally, it features a JSON format mode that ensures the model's output adheres to valid JSON syntax. This functionality is particularly beneficial for our benchmark task result collection, as we require the model to generate output in the JSON format.\footnote{\url{https://mistral.ai/news/mistral-large/}}.

\section{Prompting} A prompt serves as an input to large language models (LLMs) and plays a crucial role in determining the output of the model. Prompts generally consists of instructions, questions, input data and examples. The prompting techniques for LLMs refer to the strategies used to effectively instruct the model to generate desired outputs. This involves carefully crafting the prompt to elicit the desired response from the LLM and includes structuring the prompt, using specific phrasing, and incorporating relevant context. The pre-defined templates are very good starting point for prompts. These templates can be customized for different tasks and can ensure consistency in prompting. The effectiveness of these prompting techniques can vary depending on the specific task, the LLM being used, and the desired output. Experimentation and iterative refinement are often necessary to find the most effective prompting strategies for a given use case. Below are some of most popular prompting strategies used across wide range of use cases.
\vspace{-7pt}
\subsection{Zero-shot prompting:} Zero-shot prompting means that the prompt will not contain any examples or demonstrations. The zero-shot prompt directly instructs the model to perform a task without any additional examples to steer it. One sample prompt template used in zero-shot prompting:
\vspace{-8.5pt}
\begin{tcolorbox}[colback=white,colframe=red,listing only,listing options={basicstyle=\ttfamily}]
\textbf{Text:} "Please decide if the following text is hate speech or not:"

\textbf{Output:} "She may or may not be a Jew but she 's certainly stupid , she seems to think the Blacks won't kill her."
\end{tcolorbox}

\subsection{Few-shot prompting:} Few-shot prompting can be used as a technique to enable in-context learning where we provide examples or demonstrations in the prompt to steer the model to better performance.

\begin{tcolorbox}[colback=white,colframe=red,listing only,listing options={basicstyle=\ttfamily}]
\textbf{Text:} "Check the following job post: The title is Agile Project Manager. The location is NZ, , Wellington. The requirements is Not Mentioned. The employment type is Full-time. The industry is Not Mentioned. The function is Not Mentioned. "

\textbf{Output:} "Fake is 1, Real is 0. The job post is  0."

\textbf{Text:} "Check the following job post: The title is RN PACU . The location is US, GA, . The requirements is Not Mentioned. The employment type is Full-time. The industry is Hospital \& Health Care. The function is Not Mentioned."

\textbf{Output:} " Fake is 1, Real is 0. The job post is  1."

\textbf{Text:} "Answer the following job post: The title is Credit and Collections Clerk. The location is US, WA, . The requirements is \#NAME?. The employment type is Full-time. The industry is Hospital \& Health Care. The function is Financial Analyst."
\end{tcolorbox}

\begin{figure*}[!ht]
\centering
\includegraphics[width=\textwidth,height=12cm]{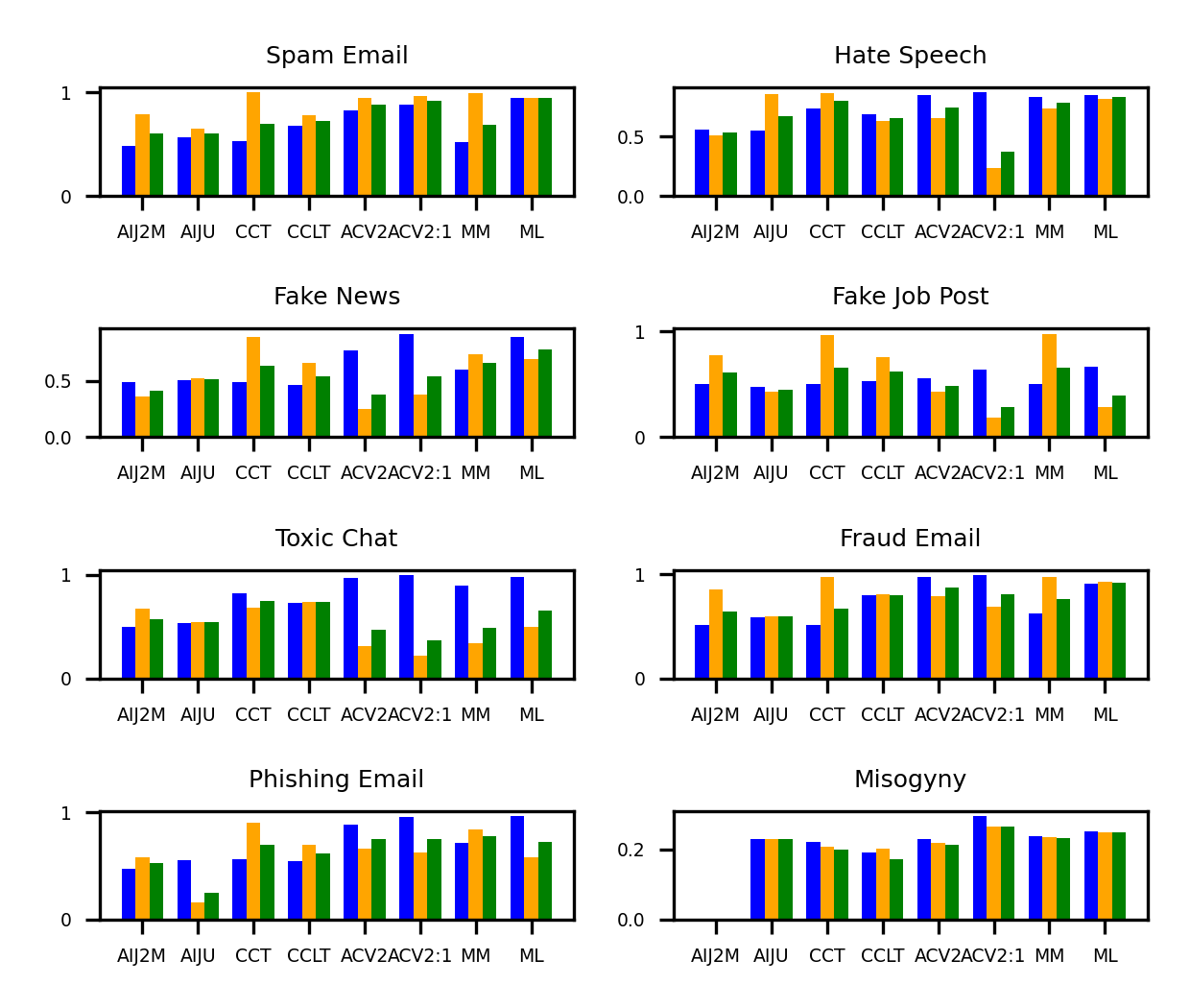}
 \caption{Classification performance of 8 LLMs for different tasks with zero shot prompting strategy. The results are shown in the form of three metrics - \textcolor{blue}{Precision (Blue)}, \textcolor{yellow}{Recall (Yellow)}, and \textcolor{green}{F1 Score (Green)}. We referred the LLMs using these abbreviations - AIJ2M - AI21.J2-Mid; AIJU - AI21.J2-Ultra; CCT - Cohere.Command-Text; CCLT - Cohere.Command-Light-Text; ACV2: Anthropic.Claude-V2; ACV2:1 - Anthropic.Claude-V2:1; MM - Mixtral MOE; ML - Mistral Large.}
\label{fig:zero_shot_figure}
\end{figure*}

\begin{figure*}[!ht]
\centering
\includegraphics[width=\textwidth,height=12cm]
{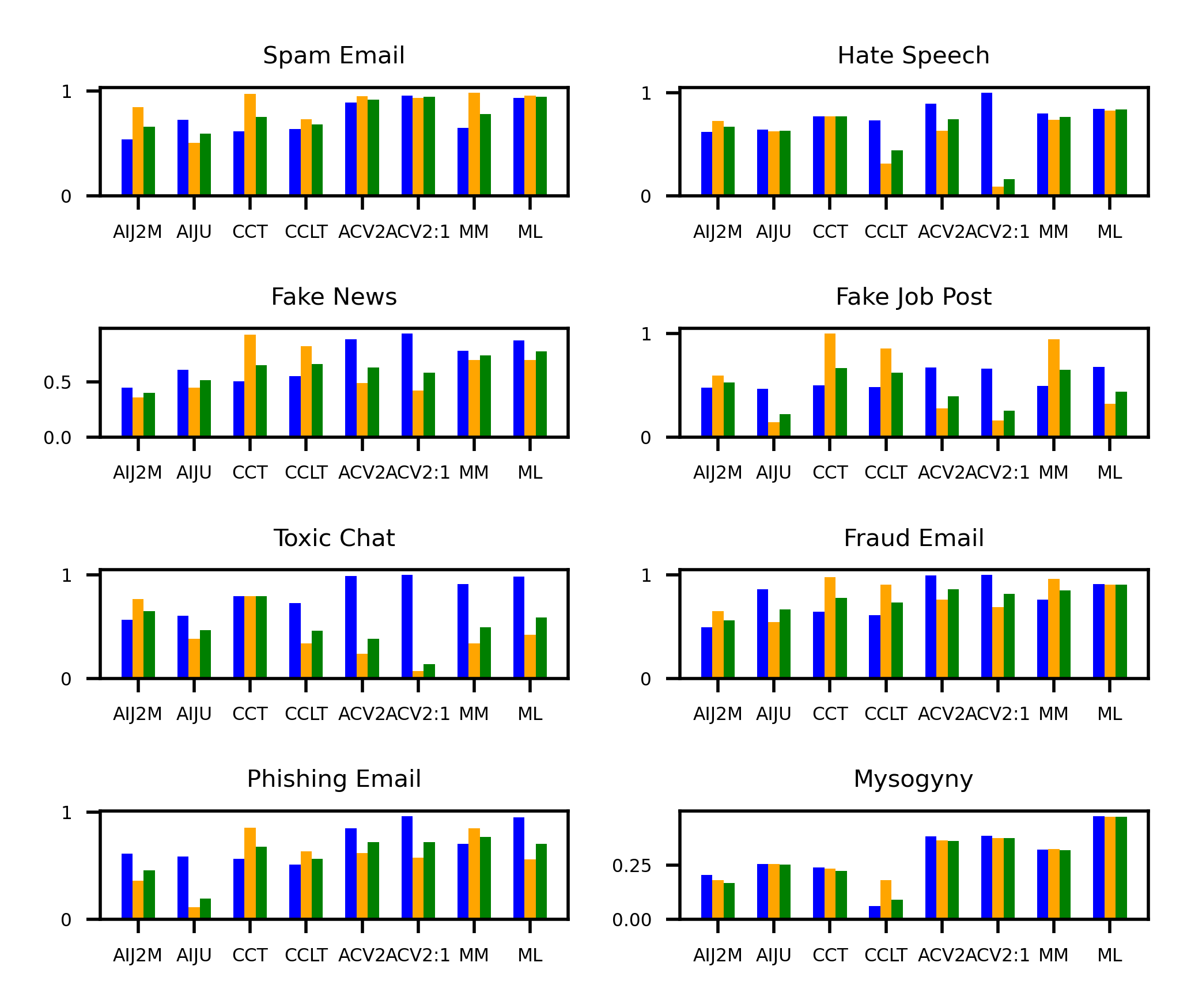}
\caption{Classification performance of 8 LLMs for different tasks with few shot prompting strategy. The results are shown in the form of three metrics - \textcolor{blue}{Precision (Blue)}, \textcolor{yellow}{Recall (Yellow)}, and \textcolor{green}{F1 Score (Green)}. }
\label{fig:few_shot_figure}
\end{figure*}

% \subsection{Chain-Of-Thought (CoT):} Introduced in Wei et al. (2022)\cite{wei2023chainofthought},Chain-of-thought (CoT) prompting enables complex reasoning capabilities through intermediate reasoning steps. We can combine it with few-shot prompting to get better results on more complex tasks that require reasoning before responding. The chain-of-thought prompting asks the model to describe the intermediate steps used to reason its way to a final answer within one response. This is useful for tasks that require detailed explanation, planning and reasoning, such as math problems and logic puzzles, where explaining the thought process is essential to fully understanding the solution. Chain-of-thought prompting aims to encapsulate the reasoning process within a single detailed, self-contained response. 

In this work, we experimented with zero-shot, and few-shot prompting. In our future iterations, we will include advanced prompting techniques such as Chain-Of-Thought, Tree of Thoughts, Prompt Chaining,  and Self-Consistency.

\begin{figure*}[!ht]
\centering
\includegraphics[width=\textwidth,height=8cm]
{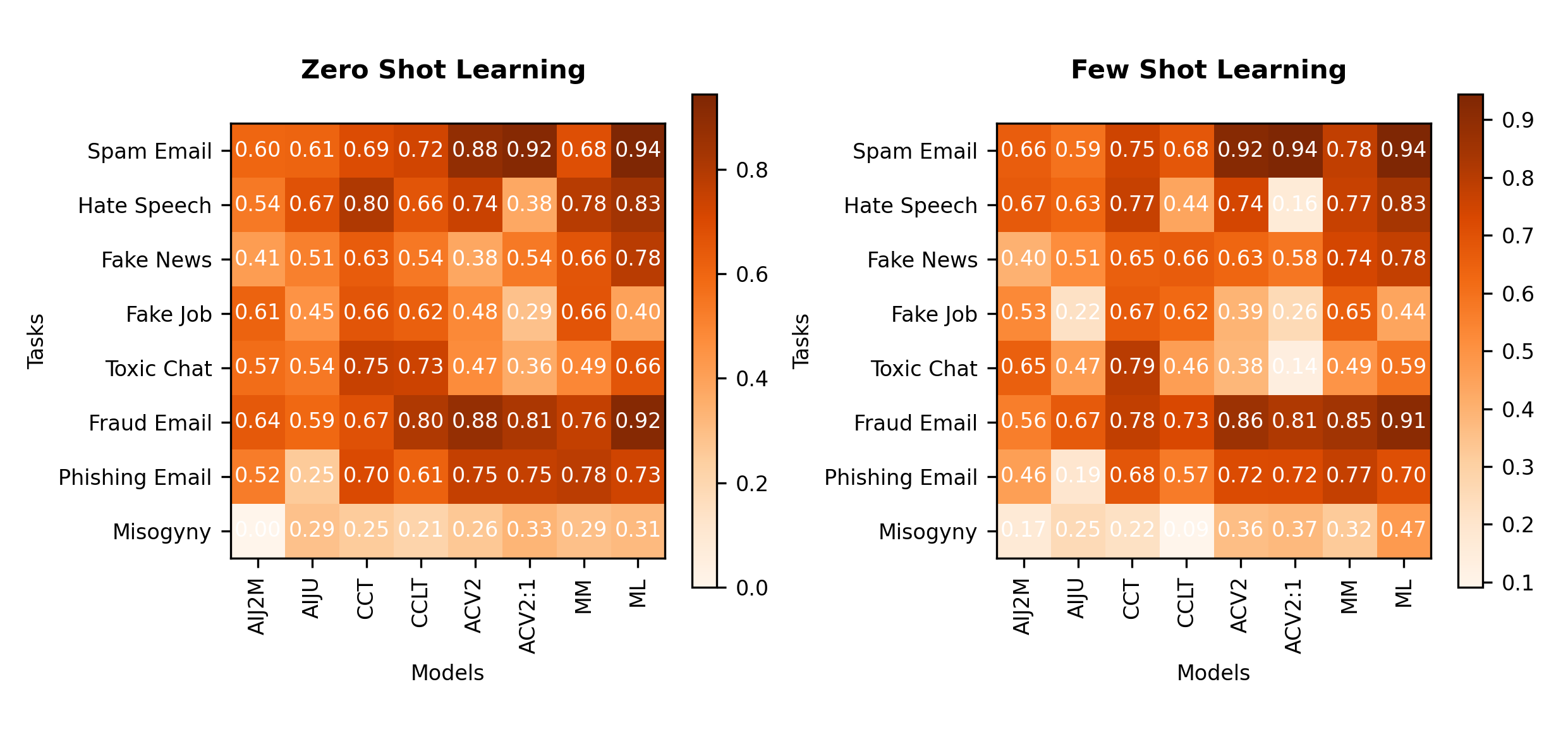}
\caption{F1 score of 8 LLMs for different tasks with zero shot prompting and few shot prompting strategy respectively.}
\label{fig:heat_map}
\end{figure*}

\section{Results}

We have experimented with eight different large language models (LLMs) across eight different datasets for classification task. For all tasks except Misogyny, the label is binary i.e. consists two categories. Like most of the fraud datasets, our datasets are also hugely imbalanced i.e. the number of abusive cases is much lower than the number of non-abusive cases. Hence, we used random under-sampling of the majority class (non-abusive) to create a balanced test set. For the Misogyny task, we transformed it into a multiple-choice problem, allowing the model to choose one correct option out of four. We tested all tasks using LLMs from the AWS Bedrock services, which provide a convenient API and a variety of LLMs for selection. To structure the output of the LLMs in a JSON format, we used LangChain parser framework\footnote{\url{https://python.langchain.com/v0.1/docs/modules/model_io/output_parsers/types/json/}}.

We have represented the classification results in form of three widely known classification metrics - \textit{precision}, \textit{recall}, and \textit{F1 score}. Figure \ref{fig:zero_shot_figure} and \ref{fig:few_shot_figure} shows precision, recall, and F1 score for zero shot and few shot prompting respectively. To get a better summary, we show the F1 score for all the experiments in the form of heat map in Figure \ref{fig:heat_map}. Here we discuss our key observations: 

\begin{itemize}
    \item \textbf{F1 Score:} For zero shot and few shot prompting, out of eight datasets, in five cases (spam email, hate speech, fake news, fraud email, misogyny) Mistral family (ML - Mistral Large) performs the best. After Mistral family, the Anthropic Claude models achieve the second best F1 score.
     
    \item \textbf{High Precision but Low Recall:} Anthropic Claude family models are highly precise in most of the cases but the recall is significantly low. For toxic chat  and hate speech detection (few-shot),  the precision is over 90\% but the recall is less than 10\%. Though these models suffer from low recall, but they can be used for proprietary use cases where highly precise results are preferred.
    
    \item \textbf{High Recall but Low Precision:} Cohere family models provide high recall i.e. can detect most of the fraud cases. However, the precision is not that great i.e. the false positive rate is high. For fake job detection (few-shot), the recall is 85\% and the precision is 48\%. For fraud email detection (few-shot), the recall is 98\% and the precision is 64\%. 

    \item \textbf{Inference Time:} AI21 family models are the fastest for inference (1.5 seconds/instance), while Mistral Large, and Anthropic Claude models are the slowest (10 seconds/instance).

    \item \textbf{Effect of Prompts:} For most of the cases, few-shot prompting does take more time for inference but does not improve the performance of the LLMs significantly. For example, for Mistral Large model, few-shot prompting improves performance in only two tasks - fake job detection, and misogyny detection. 

    \item \textbf{Format Compliance:} There are some LLMs (Command R, Command R+ from Cohere family) that could not follow the LangChain JSON parser output formats. Hence, we removed them from our final benchmark results. These models are not the best choice for production use cases as they require additional post processing of outputs which might be expensive. 

    \item \textbf{Multi-Class Classification:} In the misogyny dataset, there are five different categories. Out of five categories, four of them are different kinds of misogynistic categories (Misogynistic\_personal\_attack, Misogynistic\_pejorative, Treatment, Derogation), and one of them is non-misogynistic category.  We transformed this dataset into a multiple-choice problem, allowing the model to choose one correct option out of four. While reporting the metrics, we take the "weighted" recall, precision, and F1 across all the classes. A121 Jurassic-2 Mid (AIJ2M) model could only classify 2.1\% instances and rest of them resulted as "undecided". Similar to binary classification, here also we got the best results from Mistral, and Anthropic Claude family models. 
    
\end{itemize}

\section{Limitations}

There are several limitations of our work. 

\begin{itemize}
    \item The datasets to the best of our knowledge, are the most representative among publicly available datasets of fraud/abuse detection problems. We do not claim these datasets to be comprehensive, but hopefully with time the collection will grow to cover more business scenarios and dataset variations.

    \item While these datasets are useful for research and development of fraud detection algorithms, they do not carry any information about real fraud. If someone uses any models trained on these datasets to directly make decisions about fraud, it would cause a negative bias and could lead to false accusations.

    \item We started experimenting with relatively smaller size LLMs such as FLAN-T5, RoBERTa, mT0, etc. We decided to not include those models in our study as they require an additional  Verbalizer\footnote{\url{https://thunlp.github.io/OpenPrompt/modules/verbalizer.html}} to assign a label (abusive/non-abusive) to every textual input based on higher likelihood. On the other hand, comparatively larger LLMs that we used were capable enough to classify most of the inputs without using Verbalizer.

    \item In this study, we used zero shot and few shot prompting techniques, and followed the best practices of LLM prompting as guided by huggingface. The reason behind that is we wanted to capture the strength of the models using minimal prompt intelligence. In future version, we will also experiment with other advanced prompting techniques. 

    \item There are some advanced LLMs that we could not include in our work such as GPT family (not available on Bedrock), Llama family (Data privacy issues). The inclusion of these models could have given different set of results.

    \item This work only focuses on English language texts. We decided to use English only for two reasons. First, as most of the advanced LLMs support English and majority of prior benchmark works also focused on English only. Second, there are very few non-English textual datasets related to fraud and abuse domain that are publicly available .

\end{itemize}

\section{Dataset Disclaimers and Terms}

Firstly, the datasets used in our work were collected from the online data sources and are completely anonymous. We have carefully gone through the data and taken out anything that could have personal information in it. However, there is still a chance that some personal information might be left in the data. If anyone come across anything in the data that should not be made public please inform the authors. Secondly, these datasets may contain racism, sexuality, or other undesired content. The statements or opinions made in this dataset do not reflect the views of researchers or institutions involved in the data collection effort. Thirdly, the users of these datasets are responsible for ensuring its appropriate use and should not be utilized for training dialogue agents, or any other applications, in manners that conflict with legal and ethical standards. Finally, the users of these datasets must not attempt to determine the identity of individuals in this dataset.

\section{Conclusion \& Future Work}

In this paper, we systematically evaluate eight leading LLMs on eight different fraud and abuse categories. To the best of our knowledge, this is the first LLM benchmark specifically focused on fraud and abuse detection tasks. Based on our experiments, we make some key observations such as: a) We find larger size leads to better performance (200 Billion Anthropic family, and 176 Billion Mistral AI family are performing the best). b) Some LLMs (Mistral Large, Anthropic Claude) are much better than other LLMs (AI21, Cohere) to understand the contextual meaning for fraud and abuse classifications. c) Few-shot prompting does not always improve results over zero-shot prompting used in our experiments. For future work, we plan to experiment with fine-tuning LLM models to enhance their ability in handling fraud and abuse tasks. Additionally, in this paper, we utilized a simple LLM chain without a memory buffer. In future work, we will explore different chain structures, such as applying the Sequential Chain along with Chain-of-Thought (COT) prompts. We will also experiment with the Router Chain to process mixed types of fraud and abuse classifications.

\bibliographystyle{ACM-Reference-Format}
\bibliography{ref}

\section{Appendix}

\begin{figure*}[!h]
\centering
\includegraphics[width=0.76\textwidth, height=6.5cm]{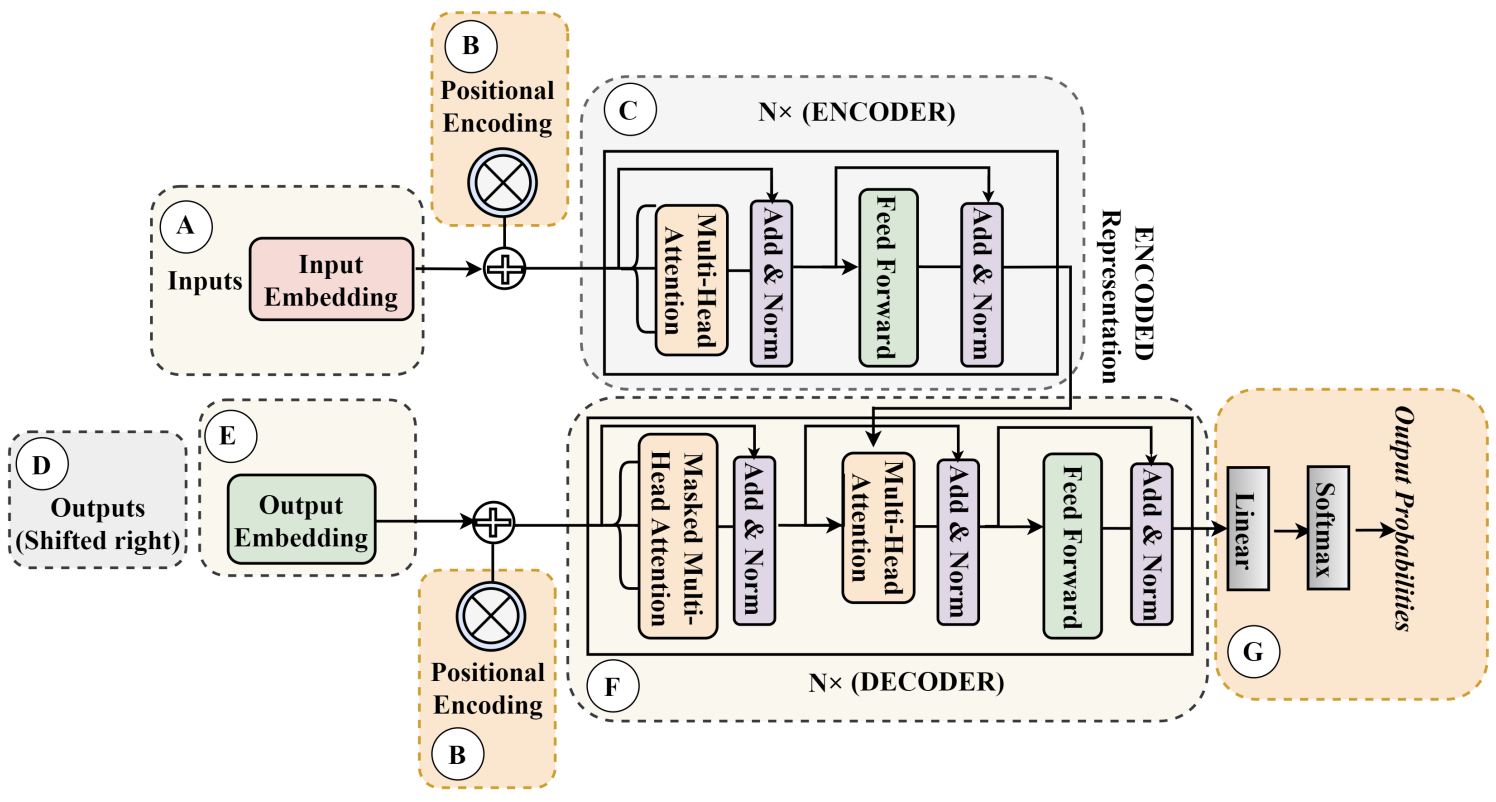}
\caption{The Transformer - model architecture~\cite{vaswani2023attention}}
\end{figure*}

\subsection{LLM Architecture}
Language modeling is a long-standing research topic, dating back to the 1950s with Shannon’s application of information theory to human language, where he measured how well simple n-gram language models predict or compress natural language text. Since then, statistical language modeling became fundamental to many natural language understanding and generation tasks, ranging from speech recognition, machine translation, to information retrieval. The recent advances on large language models (LLMs), pretrained on Web-scale text corpora, significantly extended the capabilities of language models \cite{minaee2024large}. LLMs are transformer based neural language models that contain tens to hundreds of billions of parameters which are pretrained on massive training data. The most widely used LLM architectures are encoder only\cite{devlin2019bert}, decoder only\cite{dai2019transformerxl} and encoder-decoder\cite{lewis2019bart}. 

\subsubsection{\textbf{The Transformer - model architecture:}} 
The transformer architecture was originally designed for sequence transduction or neural machine translation to convert an input sequence to an output sequence. It is a simple network architecture based solely on attention mechanisms, dispensing with recurrence and convolutions entirely\cite{vaswani2023attention}. Transformer architecture consists of seven key components. A demonstration of each of the components is shown below.

\begin{itemize}
    \item \textbf{Input Embedding:} The ML models use user-entered tokens as training data, while it can only process numeric information. Thus, it is necessary to transform these textual inputs into a numerical format known as "input embeddings". These embeddings function similarly to a dictionary, assisting the model in understanding the meaning of words by arranging them in a mathematical space where comparable phrases are situated close together.
    \item \textbf{Positional Embedding}: The order of words in a sentence is essential in the NLP field for identifying the statement’s meaning. The positional encoding is utilized to encode each word’s location in the input sequence as a collection of integers which allows the model to grasp sentence word order better and provide grammatically accurate and semantically relevant output. The original transformer architecture uses sine and cosine functions of different frequencies:
    \begin{equation}
        PE_{(pos,2i)} = sin(pos/10000^{(2i/d_{model})})
    \end{equation} 
    \begin{equation}
        PE_{(pos,2i+1)} = cos(pos/10000^{(2i/d_{model})})
    \end{equation} 
    \item \textbf{Encoder: }The encoder is composed of a stack of N = 6 identical layers. Each layer has two sub-layers.The first is a multi-headed self-attention layer and the second is a simple position-wise fully connected feed-forward network. It processes the input text and generates a series of hidden states. This consists of two linear transformations with a ReLU activation in between.
    \begin{equation}
        FFN(x) = max(0, xW_{1} + b1)W_{2} + b2
    \end{equation}
    \item \textbf{Outputs (shifted right):} During the training process, the decoder acquires the ability to predict the next word by analyzing the previous words. In this case, the output sequence is shifted by one position to the right. Consequently, the decoder is able to use the words that came before it.
    \item \textbf{Output Embedding:} The output is converted to a format known as "output embedding." Like input embeddings, output embeddings also undergo positional encoding, enabling the model to understand the order of words in a sentence.
    \item \textbf{Decoder: }The decoder is composed of a stack of N = 6 identical transformer layers. In addition to the two sub-layers in each encoder layer, the decoder has a third sub-layer, which performs multi-head attention over the output of the encoder stack. It creates output sequences from positionally encoded input sequences while learns to predict the next word from previous words in positionally encoded output embedding during the training period.
    \item \textbf{Linear Layer and Softmax: }The linear layer maps to the higher-dimensional space once the decoder has generated the output embedding. This step is required to convert the output embedding into the original input space. The softmax function generates a probability distribution for each output token in the developed vocabulary, allowing us to generate probabilistic output tokens.
\end{itemize}

% \begin{equation}
%         $MultiHead(Q, K, V) = Concat(head_{1},......,head_{h})W^o \\ 
%         where head_{i} = Attention(QW_{i}^Q, KW_{i}^K, VW_{i}^V) $
%      hhh
% \end{equation}
\begin{table*}[!h]
\caption{F1 score of 8 LLMs with 95\% confidence interval for different tasks with zero shot prompting.}
\label{zero_shot_confidence}
\begin{tabular}{lcccccccc}
               & \textbf{AIJ2M} & \textbf{AIJU} & \textbf{CCT}  & \textbf{CCLT} & \textbf{ACV2} & \textbf{ACV2:1} & \textbf{MM}   & \textbf{ML}   \\
\toprule
\textbf{Spam Email} & 0.60 $\pm$  0.03 & 0.61 $\pm$  0.03 & 0.69 $\pm$  0.03 & 0.72 $\pm$  0.05 & 0.88 $\pm$  0.02 & 0.92 $\pm$  0.02 & 0.68 $\pm$  0.03 & 0.94 $\pm$  0.01 \\
\midrule
\textbf{Hate Speech} & 0.54  $\pm$  0.04 & 0.67  $\pm$  0.03 & 0.80  $\pm$  0.03 & 0.66 $\pm$ 0.05 & 0.74 $\pm$ 0.03 & 0.38 $\pm$ 0.05 & 0.78 $\pm$ 0.03 & 0.83 $\pm$ 0.02 \\
\midrule
\textbf{Fake News} & 0.41 $\pm$ 0.05 & 0.51 $\pm$ 0.04 & 0.63 $\pm$ 0.03 & 0.54 $\pm$ 0.05  & 0.38 $\pm$ 0.05 & 0.54 $\pm$ 0.04 & 0.66 $\pm$ 0.03 & 0.78 $\pm$ 0.03 \\
\midrule
\textbf{Fake Job} & 0.61 $\pm$ 0.04 & 0.45 $\pm$ 0.04 & 0.66 $\pm$ 0.03 & 0.62 $\pm$ 0.05 & 0.48 $\pm$ 0.04 & 0.29 $\pm$ 0.05 & 0.66 $\pm$ 0.03 & 0.40 $\pm$ 0.05 \\
\midrule
\textbf{Toxic Chat} & 0.57 $\pm$ 0.04 & 0.54 $\pm$ 0.04 & 0.75 $\pm$ 0.03 & 0.73 $\pm$ 0.04 & 0.47 $\pm$ 0.05 & 0.36 $\pm$ 0.08 & 0.49 $\pm$ 0.05 & 0.66 $\pm$ 0.04 \\
\midrule
\textbf{Fraud Email} & 0.64 $\pm$ 0.04 & 0.59 $\pm$ 0.04 & 0.67 $\pm$ 0.03 & 0.80 $\pm$ 0.03 & 0.88 $\pm$ 0.02 & 0.81 $\pm$ 0.03 & 0.76 $\pm$ 0.03 & 0.92 $\pm$ 0.03 \\
\midrule
\textbf{Phishing Email} & 0.52 $\pm$ 0.05 & 0.25 $\pm$ 0.05 & 0.70 $\pm$ 0.03 & 0.61 $\pm$ 0.05 & 0.75 $\pm$ 0.03 & 0.75 $\pm$ 0.03 & 0.78 $\pm$ 0.03 & 0.73 $\pm$ 0.03 \\
\end{tabular}
\end{table*}

\begin{table*}[!h]
\caption{F1 score of 8 LLMs with 95\% confidence interval for different tasks with few shot prompting.}
\label{few_shot_confidence}
\begin{tabular}{lcccccccc}
               & \textbf{AIJ2M} & \textbf{AIJU} & \textbf{CCT}  & \textbf{CCLT} & \textbf{ACV2} & \textbf{ACV2:1} & \textbf{MM}  & \textbf{ML}   \\
\toprule
\textbf{Spam Email} & 0.66 $\pm$ 0.03 & 0.59 $\pm$  0.03 & 0.75 $\pm$  0.03 & 0.68 $\pm$  0.05 & 0.92 $\pm$  0.02 & 0.94 $\pm$  0.02 & 0.78 $\pm$  0.03 & 0.94 $\pm$  0.01 \\
\midrule
\textbf{Hate Speech} & 0.67  $\pm$  0.04 & 0.63  $\pm$  0.03 & 0.77  $\pm$  0.03 & 0.44 $\pm$ 0.05 & 0.74 $\pm$ 0.03 & 0.16 $\pm$ 0.05 & 0.77 $\pm$ 0.03 & 0.83 $\pm$ 0.02 \\
\midrule
\textbf{Fake News} & 0.40 $\pm$ 0.04 & 0.51 $\pm$ 0.04 & 0.65 $\pm$ 0.03 & 0.66 $\pm$ 0.03  & 0.63 $\pm$ 0.04 & 0.58 $\pm$ 0.04 & 0.74 $\pm$ 0.03 & 0.78 $\pm$ 0.03 \\
\midrule
\textbf{Fake Job} & 0.53 $\pm$ 0.04 & 0.22 $\pm$ 0.04 & 0.67 $\pm$ 0.03 & 0.62 $\pm$ 0.03 & 0.39 $\pm$ 0.05 & 0.26 $\pm$ 0.05 & 0.65 $\pm$ 0.03 & 0.44 $\pm$ 0.04 \\
\midrule
\textbf{Toxic Chat} & 0.65 $\pm$ 0.04 & 0.47 $\pm$ 0.04 & 0.79 $\pm$ 0.03 & 0.46 $\pm$ 0.04 & 0.38 $\pm$ 0.05 & 0.14 $\pm$ 0.08 & 0.49 $\pm$ 0.05 & 0.59 $\pm$ 0.04 \\
\midrule
\textbf{Fraud Email} & 0.56 $\pm$ 0.05 & 0.67 $\pm$ 0.04 & 0.78 $\pm$ 0.03 & 0.73 $\pm$ 0.03 & 0.86 $\pm$ 0.02 & 0.81 $\pm$ 0.03 & 0.85 $\pm$ 0.02 & 0.91 $\pm$ 0.02 \\
\midrule
\textbf{Phishing Email} & 0.46 $\pm$ 0.05 & 0.19 $\pm$ 0.05 & 0.68 $\pm$ 0.03 & 0.57 $\pm$ 0.05 & 0.72 $\pm$ 0.03 & 0.72 $\pm$ 0.03 & 0.77 $\pm$ 0.03 & 0.70 $\pm$ 0.03 \\
\end{tabular}
\end{table*}

\subsubsection{\textbf{Transformer: Encoder}} Encoder models use only the encoder of a Transformer model \footnote{ 
\url{https://huggingface.co/learn/nlp-course/en/chapter1/5}}. At each stage, the attention layers can access all the words in the initial sentence. These models are often characterized as having “bi-directional” attention, and are often called auto-encoding models. The pretraining of these models usually revolves around somehow corrupting a given sentence (for instance, by masking random words in it) and tasking the model with finding or reconstructing the initial sentence. Encoder models are best suited for tasks requiring an understanding of the full sentence, such as sentence classification, named entity recognition (and more generally word classification), and extractive question answering. The representatives of this family of include: ALBERT\cite{lan2020albert}, BERT\cite{devlin2019bert}, DistilBERT\cite{sanh2020distilbert}, ELECTRA \footnote{\url{https://openreview.net/pdf?id=r1xMH1BtvB}}, RoBERTa\cite{liu2019roberta}. 

% \subsubsection{BERT:}One of the prominent encoder only model is BERT (Bi-directional Encoder Representations from Transformers)\cite{devlin2019bert}. BERT is an encoder-only Transformer that randomly masks certain tokens in the input to avoid seeing other tokens, which would allow it to “cheat”. The pretraining objective is to predict the masked token based on the context. This allows BERT to fully use the left and right contexts to help it learn a deeper and richer representation of the inputs. \textcolor{red}{<<REMOVE>>}

% \subsubsection{RoBERTa:}RoBERTa\cite{liu2019roberta} improved upon this by introducing a new pretraining recipe that includes training for longer and on larger batches, randomly masking tokens at each epoch instead of just once during preprocessing, and removing the next-sentence prediction objective. The dominant strategy to improve performance is to increase the model size. But training large models is computationally expensive. One way to reduce computational costs is using a smaller model like DistilBERT. \textcolor{red}{<<REMOVE>>}

\subsubsection{\textbf{Transformer: Decoder}} Decoder models use only the decoder of a Transformer model\footnote{\url{https://huggingface.co/learn/nlp-course/en/chapter1/6}}. At each stage, for a given word the attention layers can only access the words positioned before it in the sentence. These models are often called auto-regressive models. The pretraining of decoder models usually revolves around predicting the next word in the sentence. These models are best suited for tasks involving text generation. The representatives of this family of include: CTRL\cite{keskar2019ctrl}, GPT \cite{Radford2018ImprovingLU}, GPT-2 \cite{Radford2019LanguageMA}, Transformer XL\cite{dai2019transformerxl}, Llama 2\cite{touvron2023llama}. 

% \subsubsection{Llama 2:} Llama 2\cite{touvron2023llama}, an updated version of Llama 1, trained on a new mix of publicly available data. The Llama 2 release introduces a family of pretrained and fine-tuned LLMs, ranging in scale from 7B to 70B parameters (7B, 13B, 70B). The pretrained models come with significant improvements over the Llama 1 models, including being trained on 40\% more tokens, having a much longer context length (4k tokens), and using grouped-query attention for fast inference of the 70B model. \textcolor{red}{<<REMOVE>>}

\subsubsection{\textbf{Transfomer: Encoder-Decoder:}} Encoder-decoder models (also called sequence-to-sequence models) use both parts of the Transformer architecture\footnote{\url{https://huggingface.co/learn/nlp-course/en/chapter1/7?fw=pt}}. At each stage, the attention layers of the encoder can access all the words in the initial sentence, whereas the attention layers of the decoder only access the words positioned before a given word in the input. The pretraining of these models can be done using the objectives of encoder or decoder models, but usually involves something a bit more complex. For instance, some models are pretrained by replacing random spans of text that can contain several words with a single mask special word, and the objective is then to predict the text that this mask word replaces. Sequence-to-sequence models are best suited for tasks revolving around generating new sentences depending on a given input, such as summarization, translation, or generative question answering. The representatives of this family of models include: BART\cite{lewis2019bart}, mBART\cite{liu2020multilingual}, Marian\footnote{\url{https://huggingface.co/docs/transformers/model_doc/marian}}, T5\cite{raffel2023exploring}

\subsubsection{\textbf{Instruction Fine-tuned Models:}}Instruction tuning is a simple method that combines appealing aspects of both the pretrain–finetune and prompting paradigms by using supervision via finetuning to improve the ability of language models to respond to inference-time text interactions. The supervision teaches the model to perform tasks described via instructions. Recent empirical results\cite{wei2022finetuned}demonstrate promising abilities of language models to perform tasks described purely via instructions. Finetuning on groups of language tasks has been shown to significantly boost this zero-shot task generalization of language models \cite{wei2022finetuned, chung2022scaling}.

% \subsubsection{FLAN-T5:}Flan-T5 is an instruction fine-tuned version of T5 or Text-to-Text Transfer Transformer Language Model \cite{raffel2023exploring}. It is a dense encoder-decoder model based on pretrained T5 and fine-tuned with instructions for better zero-shot and few-shot performance. There are 5 different sizes of this model: (1) Flan-T5-Small (80M Parameters), (2) Flan-T5-Base (250M Parameters), (3) Flan-T5-Large (780M Parameters), (4) Flan-T5-XL (3B Parameters) and (5) Flan-T5-XXL (11B Parameters).The finetuning data comprises 473 datasets, 146 task categories, and 1,836 total tasks which includes toxic language detection. \textcolor{red}{<<REMOVE>>}

% \subsubsection{mT0:} mT0 is a Multitask prompted finetuning (MTF) variant of mT5\cite{muennighoff2023crosslingual}. The model is fine-tuned using BigScience's crosslingual task mixture i.e xP3 datasets, creating a variety of new models capable of crosslingual generalization to unseen tasks and languages.There are 5 different sizes of this model: (1) mT0-Small (300M Parameters), (2) mT0-Base (580M Parameters), (3) mT0-Large (1.2B Parameters), (4) mT0-XL (3.7B Parameters) and (5) mT0-XXL (13B Parameters). \textcolor{red}{<<REMOVE>>}

\subsection{LangChain Framework}
In this paper, we leverage LangChain for in-context learning. LangChain is a robust Python library designed to simplify interactions with various LLM providers. Chains are a vital component of LangChain, allowing multiple elements to seamlessly integrate. In this work, we specifically used the LLMChain\footnote{\url{https://api.python.langchain.com/en/latest/chains/langchain.chains.llm.LLMChain.html}}.

\subsection{Additional Results}
We have reported precision, recall, and F1 score for various experiments in the Results section above. In Table \ref{zero_shot_confidence}, and Table \ref{few_shot_confidence}, we are showing the 95\% confidence interval along with the F1 score for different tasks and different models. The key observation here is the variance is low that means LLMs are stable in abuse detection. 

\end{document}